%% file: token_efficient_rl_arxiv.tex
\documentclass[dvipsnames]{article} %
\usepackage{iclr2026_conference,times}

\input{math_commands.tex}

\usepackage{wrapfig}
\usepackage{natbib}
\usepackage{hyperref}
\usepackage{url}
\usepackage{graphicx}
\usepackage{amsthm}
\usepackage{amssymb}
\usepackage{tcolorbox}
\usepackage{enumitem}
\usepackage{subcaption}
\usepackage{dsfont}
\usepackage{booktabs}
\usepackage{listings}
\usepackage{makecell} %
\usepackage{subcaption}
\usepackage{caption}
\usepackage{siunitx}
\usepackage[normalem]{ulem}
\usepackage{textcomp} 

\usepackage{algorithmic}
\usepackage{algorithm}

\usepackage{siunitx}
\usepackage[table]{xcolor}

\usepackage{xcolor}
\definecolor{codebg}{RGB}{248,248,248}

\definecolor{goodbg}{RGB}{232,245,233} 
\definecolor{badbg}{RGB}{255,235,238}  
\definecolor{modelbg}{RGB}{220,230,250}
\definecolor{modelbgA}{RGB}{220,230,250} 
\definecolor{modelbgB}{RGB}{235,245,235} 

\definecolor{ciGood}{RGB}{0,128,0}
\definecolor{ciBad}{RGB}{180,0,0}
\definecolor{ciGrey}{RGB}{120,120,120}
\newcommand{\cig}[1]{\textcolor{ciGood}{#1}} 
\newcommand{\cib}[1]{\textcolor{ciBad}{#1}}  
\newcommand{\cir}[1]{\textcolor{ciGrey}{#1}}
\usepackage[cache=false]{minted} %
\usepackage{booktabs}
\usepackage{siunitx}

\renewcommand{\cite}{\citep}
\usepackage{amsthm}
\newtheorem{theorem}{Theorem}
\newtheorem{proposition}[theorem]{Proposition}
\title{Not All Tokens are Needed: Token-Efficient Reinforcement Learning}

\makeatletter
\renewcommand*{\thefootnote}{\fnsymbol{footnote}} 
\makeatother

\author{\footnotesize \normalfont
  \begin{tabular}[t]{@{}l@{}}
  \textbf{Hejian Sang}\textsuperscript{1\,*\,\textdagger}\quad
  \textbf{Yuanda Xu}\textsuperscript{1\,*}\quad
  \textbf{Zhengze Zhou}\textsuperscript{1\,*}\quad
  \textbf{Ran He}\textsuperscript{1\,*}\quad
  \textbf{Zhipeng Wang}\textsuperscript{1}
  \end{tabular}
  \\[-1pt]
  \footnotesize \textsuperscript{1} LinkedIn Corporation, CA, USA%
}

\iclrfinalcopy
\begin{document}

\maketitle

\begingroup
\renewcommand\thefootnote{\fnsymbol{footnote}}
\setcounter{footnote}{0}%
\footnotetext[1]{Equal contribution.}
\footnotetext[2]{\mbox{Corresponding author: Hejian Sang\textless{}\texttt{hsang@linkedin.com}\textgreater{}}}
\endgroup

\begin{abstract}
Reinforcement learning (RL) has become a key driver of progress in large language models, but scaling RL to long chain-of-thought (COT) trajectories remains increasingly constrained by \emph{backpropagation over every generated token}. Even with optimized rollout engines, full-token updates can consume a large fraction of total training cost, turning token length into a hidden tax on RL. 
We introduce \textbf{Not All Tokens are Needed (NAT)}, a unified framework that makes the \emph{token budget} a first-class optimization primitive: NAT updates the policy using only a selected subset of generated tokens while preserving the learning signal of full-sequence RL. 
The core idea is an unbiased partial-token policy-gradient estimator via \emph{Horvitz--Thompson} reweighting, which ensures statistically correct gradients despite subsampling. 
We instantiate NAT with two simple, plug-and-play token selection schemes: \textit{Uniform Random Sampling (URS)} and \textit{Random Prefix Cutting (RPC)}, both of which reduce forward/backward compute and memory without modifying the reward computation or rollout pipeline.
Across mathematical reasoning benchmarks, NAT matches full-token GRPO performance while using as few as \textbf{50\%} of tokens, providing an efficient and orthogonal pathway to scaling RL beyond the limits imposed by long trajectories. From our experiments, \textit{RPC} can save \textbf{18}\% peak GPU memory and \textbf{29}\% forward and backward RL training time for Qwen3-8B RL training.
\end{abstract}

\begin{figure}[H]
\centering
  \includegraphics[width=\textwidth]{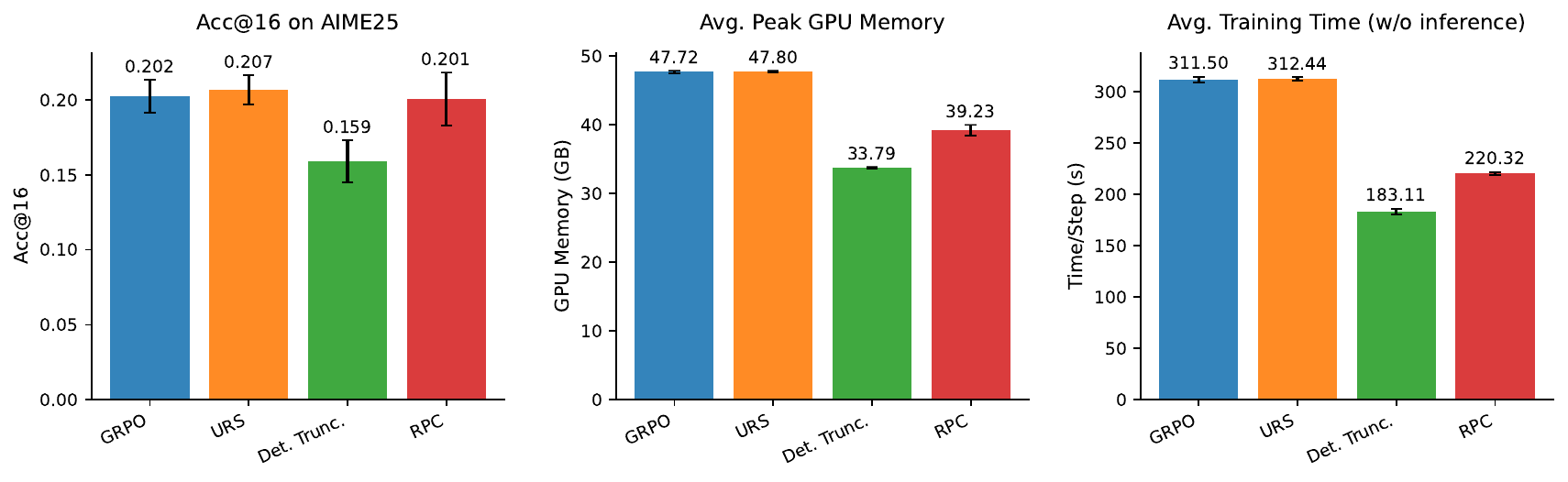}
  \caption{Barplots of Qwen3-8B RL training metrics with 95\% CIs across 5 runs for \textbf{GRPO} (vanilla GRPO), \textbf{URS} (GRPO with random sampling $p=0.5$), \textbf{Det. Trunc.} (GRPO with deterministic prefix truncation of 50\% tokens) and \textbf{RPC} (GRPO with uniform random prefix cutting).}
  \label{fig:teaser}
\end{figure}

\input{1_Introduction}

\input{2_Preliminaries}

\input{3_Token_RL}

\input{4_RPC}

\input{5_Exps}

\input{6_Related_Work}

\input{7_Limitation}

\input{8_Discussion}


\bibliographystyle{iclr2026_conference}

\newpage

\appendix
\input{9_Appendix}

\end{document}

%% file: math_commands.tex
\usepackage{amsmath,amsfonts,bm}

\def\eqref#1{equation~\ref{#1}}

\def\1{\bm{1}}

\DeclareMathAlphabet{\mathsfit}{\encodingdefault}{\sfdefault}{m}{sl}
\SetMathAlphabet{\mathsfit}{bold}{\encodingdefault}{\sfdefault}{bx}{n}

%% file: 1_Introduction.tex
\section{Introduction}

Recent generations of Large Language Models (LLMs) \cite{ziegler2019fine, ouyang2022training, stiennon2020learning} have demonstrated striking advances in mathematical reasoning \cite{guo2025deepseek}, code synthesis \cite{pan2024training}, and scientific problem solving. A major driver behind these gains is post-training with reinforcement learning objectives that use \emph{verifiable} signals (like unit tests for code, symbolic execution, or exact-answer checks for math) commonly referred to as Reinforcement Learning from Verifiable Rewards (RLVR) \cite{jaech2024openai, guo2025deepseek, team2025kimi, yang2025qwen3}. By directly optimizing for task success, RLVR can push models beyond what supervised fine-tuning alone achieves, especially on benchmarks where solutions require multi-step chains-of-thought (CoT) and careful intermediate decisions. As a result, RLVR has become a core ingredient in modern reasoning systems, including OpenAI O1 \cite{jaech2024openai}, DeepSeek-R1 \cite{guo2025deepseek}, Kimi 1.5 \cite{team2025kimi}, and Qwen3 \cite{yang2025qwen3}.

Despite its effectiveness, RLVR remains expensive and increasingly difficult to scale. A central reason is that standard RLVR pipelines treat \emph{all} generated tokens as equally important for learning: they compute policy-gradient losses and backpropagate through every token in each rollout trajectory. However, as RLVR succeeds at eliciting deeper reasoning, trajectories often become longer and more variable, which amplifies activation memory, increases per-update FLOPs, and can trigger out-of-memory (OOM) failures or unstable optimization dynamics \cite{guo2025deepseek, yeo2025demystifying}. In practice, the cost of RLVR grows not only with the number of rollouts, but also with the length of each rollout and long-CoT behavior can become the training bottleneck.

A typical GRPO \cite{shao2024deepseekmath} training cycle further highlights an efficiency mismatch across stages. Rollout generation is often the most visible bottleneck and can account for a large fraction of wall-clock time (e.g., around 70\% in representative profiles \cite{gao2025rollpacker}). Consequently, recent work has aggressively optimized generation through high-throughput inference engines \citep{kwon2023efficient,zheng2024sglang}, speculative decoding \citep{leviathan2023fast} and low precision inference engine \cite{xi2026jet}. Yet, even when generation is accelerated, the subsequent \emph{learning} phase—forward and backward passes over full trajectories—remains memory-bound and compute-intensive. This creates a practical ceiling: we can generate rollouts faster, but we cannot \emph{consume} them efficiently enough to translate that throughput into scalable policy improvement.

This paper takes an orthogonal approach to RLVR efficiency. Instead of optimizing how rollouts are produced, we optimize how they are \emph{used} by the learner. We ask a simple but consequential question:
\begin{quote}
\noindent\textbf{Do we really need all tokens to train strong RL reasoners?}
\end{quote}
Intuitively, not all tokens in a long CoT contribute equally to the learning signal. Many tokens are “mechanical” continuation (for example, filling in algebraic steps) reiterating previously established context, or emitting low-entropy boilerplate, while a smaller subset corresponds to high-impact decision points that steer the trajectory. If we can update the policy using only a carefully chosen subset of tokens—without distorting the training objective, we can reduce memory and compute per optimizer update, increase training throughput, and unlock longer-context RLVR without sacrificing performance.

We introduce \emph{Not All Tokens are Needed} (NAT), a principled framework for token-efficient RLVR. NAT performs policy optimization using only a selected subset of tokens from each rollout while keeping reward signals computed on the \emph{full} response. Concretely, NAT masks tokens during the policy-gradient computation, so that only included tokens contribute to backpropagation. This truncates the activation graph, reduces memory pressure, and decreases per-step compute. Importantly, NAT is not restricted to one masking rule: it can instantiate uniform token sampling, structured prefix-based schemes, or other selection mechanisms, as long as we can correct for selection bias appropriately.

The core technical ingredient of NAT is \textbf{Horvitz--Thompson (HT) estimation} applied to token-masked policy gradients. Each token $t$ is assigned a inclusion probability, and NAT reweights gradients by the inverse of that probability. We prove that this HT-corrected token-masked objective yields an \emph{unbiased} estimator of the original full-sequence RLVR gradient. In other words, NAT preserves the target learning signal in expectation, even though each update only backpropagates through a subset of tokens. This provides a rigorous foundation for token-efficient RLVR: partial-token updates can remain statistically aligned with optimizing the full-sequence reward objective, rather than introducing systematic drift.

We instantiate NAT on top of the GRPO objective and evaluate across multiple reasoning benchmarks and model scales. Empirically, NAT achieves performance comparable to full-token GRPO while using substantially fewer tokens in the backward pass (e.g., as low as 50\% token coverage), translating into lower activation memory and higher training throughput. Moreover, NAT composes naturally with system-level optimizations (faster rollout engines, batching, scheduling), because it targets a different bottleneck: the cost of consuming long trajectories during learning.

\paragraph{Contributions.}
Our main contributions are:
\begin{itemize}
    \item \textbf{A unified framework for token-efficient RLVR.} We propose \emph{Not All Tokens are Needed} (NAT), which performs RLVR updates using masked token subsets while preserving full-response reward evaluation. NAT supports a broad family of token selection strategies, including uniform random token sampling and structured prefix-based masking.
    \item \textbf{Unbiasedness via Horvitz--Thompson correction.} We show that HT reweighting yields an unbiased estimator of the full-token GRPO gradient for any positive inclusion probabilities. This establishes a principled connection between token masking and faithful optimization of the original RLVR objective.
    \item \textbf{Practical speedups without sacrificing reasoning quality.} Through experiments on math reasoning benchmarks, we demonstrate that NAT—especially \emph{Random Prefix Cutting (RPC)}—can match full-token GRPO while reducing backpropagation volume by up to \textbf{50\%}, yielding direct savings: \textbf{18}\% less peak GPU memory and \textbf{29}\% less forward and backward RL training time for Qwen3-8B RL training using DAPO dataset \cite{yu2025dapo}.
\end{itemize}

%% file: 2_Preliminaries.tex
\section{Preliminaries}
\label{sec:prelim}

\subsection{Notation and RL Setup}

We consider an autoregressive policy $\pi_\theta$ that generates a response
$o=(o_1,\ldots,o_{T})$ conditioned on a prompt $q \sim \mathcal{D}$.
The policy factorizes as
$
\pi_\theta(o \mid q)=\prod_{t=1}^{T}\pi_\theta(o_t \mid q,o_{<t}).
$
In RL with verifiable rewards (RLVR), each completed response $o$ receives a
scalar reward $R(q,o)$ computed by an automatic verifier (e.g., exact-match,
unit tests, or rubric-based scoring).

We distinguish the \emph{current} policy $\pi_\theta$ (being optimized) from the
\emph{behavior} policy $\pi_{\theta_{\text{old}}}$ used to collect rollouts.
For each prompt $q$, we sample a group of $G$ responses
$\{o_i\}_{i=1}^{G} \sim \pi_{\theta_{\text{old}}}(\cdot\mid q)$ and compute rewards
$\{R_i\}_{i=1}^{G}$, where $R_i := R(q,o_i)$.

For token $t$ in response $i$, we define the standard importance ratio
\begin{equation}
r_{i,t}(\theta)
=
\frac{\pi_\theta(o_{i,t}\mid q,o_{i,<t})}
{\pi_{\theta_{\text{old}}}(o_{i,t}\mid q,o_{i,<t})}.
\label{eq:ratio}
\end{equation}
This ratio corrects for the fact that data are collected under $\pi_{\theta_{\text{old}}}$
while we optimize $\pi_\theta$.

\subsection{Group Relative Policy Optimization (GRPO)}
\label{subsec:grpo}

Group Relative Policy Optimization (GRPO) is a PPO-style policy-gradient method
that removes the learned value function (critic) and instead constructs a baseline
directly from the \emph{group} of sampled responses for each prompt. Intuitively,
responses in the same group compete against each other: a response is reinforced
only insofar as it is better than its peers. This provides a low-variance advantage
signal without training a separate critic model.  

\paragraph{Group-relative advantage.}
Given rewards $\{R_i\}_{i=1}^{G}$ for prompt $q$, GRPO defines a normalized
advantage per response:
\begin{equation}
\hat{A}_i
=
\frac{R_i - \mu_R}{\sigma_R + \epsilon},
\qquad
\mu_R=\frac{1}{G}\sum_{j=1}^{G}R_j,
\quad
\sigma_R=\sqrt{\frac{1}{G}\sum_{j=1}^{G}(R_j-\mu_R)^2},
\label{eq:grpo-adv}
\end{equation}
where $\epsilon>0$ is a small constant for numerical stability.
GRPO then shares this response-level advantage across tokens:
$\hat{A}_{i,t} := \hat{A}_i$ for all $t$.
This design matches the RLVR setting where supervision is sequence-level
(correct/incorrect), so all tokens in a completion receive the same global credit.

\paragraph{Clipped surrogate objective.}
Like PPO \cite{schulman2017proximal}, GRPO uses a clipped importance-weighted surrogate to prevent overly
large policy updates. For token $t$ in response $o_i$, define the PPO-style
clipped surrogate
\begin{equation}
S_{i,t}(\theta)
=
\min\Big(
r_{i,t}(\theta)\,\hat{A}_{i,t},\;
\mathrm{clip}\big(r_{i,t}(\theta), 1-\varepsilon, 1+\varepsilon\big)\,\hat{A}_{i,t}
\Big),
\label{eq:grpo-surrogate}
\end{equation}
where $\varepsilon$ is the clip threshold. The $\min(\cdot)$ yields a conservative
(pessimistic) objective: once $r_{i,t}(\theta)$ leaves the trust region
$[1-\varepsilon,\,1+\varepsilon]$, further probability changes no longer increase
the surrogate, which stabilizes optimization.

\paragraph{Token-level GRPO loss with KL regularization.}
To prevent the policy from drifting too far from a fixed reference policy
$\pi_{\text{ref}}$ (e.g., the SFT model), we incorporate a KL penalty directly
into the token-level objective. We define the per-token GRPO loss as
\begin{equation}
L_{i,t}^{\text{GRPO}}(\theta)
=
S_{i,t}(\theta)
-
\beta\,D_{\mathrm{KL}}\!\big(\pi_\theta \,\|\, \pi_{\text{ref}}\big),
\label{eq:grpo-token-loss}
\end{equation}
where $D_{\mathrm{KL}}(\pi_\theta \,\|\, \pi_{\text{ref}})$ denotes a per-token KL
term evaluated at the same conditioning context $(q,o_{i,<t})$, and $\beta\ge 0$
controls the strength of regularization.\footnote{Many implementations approximate
this KL using the sampled-action log-prob difference
$\log\pi_\theta(o_{i,t}\mid q,o_{i,<t})-\log\pi_{\text{ref}}(o_{i,t}\mid q,o_{i,<t})$;
we keep the notation abstract for clarity.} Although we present the general GRPO formulation with KL regularization for completeness, recent work advocates for removing KL divergence \cite{yu2025dapo}.

\paragraph{Sequence-level aggregation.}
GRPO aggregates the token-level loss by averaging over tokens within each response
and then over responses in the group:
\begin{equation}
J_{\text{GRPO}}(\theta)
=
\mathbb{E}_{q\sim\mathcal{D},\,\{o_i\}\sim\pi_{\theta_{\text{old}}}}
\Bigg[
\frac{1}{G}\sum_{i=1}^{G}\frac{1}{|o_i|}\sum_{t=1}^{|o_i|}
L_{i,t}^{\text{GRPO}}(\theta)
\Bigg].
\label{eq:grpo-obj}
\end{equation}

\subsection{GRPO Training Pipeline}
\label{subsec:grpo-pipeline}

A GRPO iteration can be viewed as a three-stage pipeline with distinct system
bottlenecks:

\paragraph{(1) Rollout / Inference.}
For each prompt $q$, we sample a group of $G$ completions $\{o_i\}$ from the
behavior policy $\pi_{\theta_{\text{old}}}$ (a lagged snapshot of the current policy).
This stage is throughput-critical and is often served by high-performance inference
engines (e.g., \texttt{vLLM} \cite{kwon2023efficient} or \texttt{SGLang} \cite{zheng2024sglang} ) that support continuous batching and
memory-efficient KV-cache management.

\paragraph{(2) Forward pass (scoring + log-probs).}
Given the collected trajectories, we evaluate each completion with the verifier to
obtain rewards $\{R_i\}$ and compute group-relative advantages $\{\hat A_i\}$ via
Eq.~\eqref{eq:grpo-adv}. In parallel, we run a forward pass of the current policy
$\pi_\theta$ to compute token log-probabilities
$\log\pi_\theta(o_{i,t}\mid q,o_{i,<t})$, which are needed to form the importance
ratios $r_{i,t}(\theta)$. Since reward evaluation is typically much cheaper than the
model forward pass and can be overlapped with it, we refer to this combined stage
as the \emph{forward pass} for simplicity.

\paragraph{(3) Backward pass (optimization).}
We backpropagate through the GRPO objective (Eq.~\eqref{eq:grpo-obj}) and update
parameters using an optimizer such as AdamW \cite{loshchilov2017decoupled}. This
stage is often memory-dominated because it must store activations for
backpropagation over long chain-of-thought trajectories.

This separation is useful in practice: rollout favors inference-optimized kernels
and scheduling, while the training stages are dominated by activation memory and
stable optimization.

%% file: 3_Token_RL.tex
\section{Token-Efficient Reinforcement Learning}
\label{sec:token-efficient-rl}

RLVR pipelines decouple \emph{reward evaluation} from \emph{optimization}: rewards must be
computed on the \emph{full} generated response to faithfully reflect task success,
but the policy-gradient update is typically formed by summing token-level log-prob
terms over the entire trajectory. This creates a major efficiency
opportunity. In particular, long chain-of-thought (CoT) trajectories inflate the
activation footprint and FLOPs of the training stage, since standard GRPO updates
backpropagate through \emph{every} token.

We propose a general framework that reduces the \emph{optimization} cost by
performing policy updates on a \emph{subset} of tokens, while still using the
\emph{full response} to compute rewards and advantages. The key technical tool is
Horvitz--Thompson (HT) estimation, which provides unbiased estimators of
full-sequence objectives under randomized token selection. Conceptually, we keep
the RL signal (\(R_i\), \(\hat A_i\)) unchanged, and only sparsify which tokens
contribute to the gradient.

\subsection{General Framework: Token Masking with Horvitz--Thompson Reweighting}
\label{subsec:token-masking-ht}

Consider a prompt \(q\) and a sampled response \(o_i=(o_{i,1},\dots,o_{i,T_i})\)
of length \(T_i\). Recall from Section~\ref{sec:prelim} that GRPO defines a
token-level loss \(L^{\text{GRPO}}_{i,t}(\theta)\) and optimizes the sequence-average objective by
averaging over \(t=1,\dots,T_i\) and over samples \(i=1,\dots,G\).

\paragraph{Random masking.}
We introduce a binary mask \(m_{i,t}\in\{0,1\}\) indicating whether token \(t\) of
trajectory \(i\) participates in the policy update:
\begin{equation*}
m_{i,t} \sim \mathrm{Bernoulli}(p_{i,t}),\qquad p_{i,t}\in(0,1].
\label{eq:general-mask}
\end{equation*}
Here \(p_{i,t}\) is the inclusion probability.

\paragraph{HT estimator for the per-sequence mean loss.}
Define the full-token per-sequence mean loss
\(\mu_i(\theta) = \frac{1}{T_i}\sum_{t=1}^{T_i} L^{\text{GRPO}}_{i,t}(\theta)\).
Using only the selected tokens, the Horvitz--Thompson estimator is
\begin{equation}
\widehat{\mu}_i^{\text{HT}}(\theta)
\;=\;
\frac{1}{T_i}\sum_{t=1}^{T_i}\frac{m_{i,t}}{p_{i,t}}\,
L^{\text{GRPO}}_{i,t}(\theta).
\label{eq:ht-estimator}
\end{equation}

\begin{proposition}[Unbiasedness of HT token masking]
\label{prop:ht-unbiased}
For any inclusion probabilities \(\{p_{i,t}\}_{t=1}^{T_i}\) with \(p_{i,t}>0\),
\(\mathbb{E}_m[\widehat{\mu}_i^{\text{HT}}(\theta)] = \mu_i(\theta)\).
Moreover, under standard regularity conditions allowing interchange of gradient
and expectation, \(\mathbb{E}_m[\nabla_\theta \widehat{\mu}_i^{\text{HT}}(\theta)]
= \nabla_\theta \mu_i(\theta)\). 
\end{proposition}

\noindent
The proof is shown in Appendix \ref{app:gradient-unbias}.
Consequently, replacing the full-token GRPO
objective with the masked objective formed by \(\widehat{\mu}_i^{\text{HT}}\) yields
an unbiased estimator of the full-token GRPO gradient. The inclusion probabilities \(p_{i,t}\) define a compute--variance trade-off:
smaller \(p_{i,t}\) reduces the expected number of backpropagated tokens, but
increases estimator variance through the HT weights \(1/p_{i,t}\).
Next we instantiate this framework with two practical masking schemes.

\paragraph{Uniform Random Token Sampling (URS)}
URS sets a constant inclusion probability \(p_{i,t}=p\) for all \((i,t)\).
Intuitively, URS keeps a fraction \(p\) of tokens on average, providing
approximately linear savings in the backward pass. Under HT correction,
the per-token score-function contribution scales as \(1/p\).
For example, for a generic score term
\(g_{i,t}(\theta) = \nabla_\theta \log \pi_\theta(o_{i,t}\mid q,o_{i,<t})\),
the HT-corrected contribution becomes
\begin{equation*}
\widehat{g}^{\textsc{urs}}_{i,t}(\theta)
\;=\;
\frac{m_{i,t}}{p}\, g_{i,t}(\theta),
\qquad m_{i,t}\sim\mathrm{Bernoulli}(p).
\label{eq:urs-ht-score}
\end{equation*}
This is unbiased: \(\mathbb{E}_m[\widehat{g}^{\textsc{urs}}_{i,t}(\theta)] = g_{i,t}(\theta)\).

\paragraph{Gradient-norm inflation under URS}
While URS is unbiased, HT reweighting inflates the second moment by \(1/p\):
\begin{align*}
\mathbb{E}_m\!\left[\left\|\widehat{g}^{\textsc{urs}}_{i,t}(\theta)\right\|^2\right]
&=
\mathbb{E}_m\!\left[\left\|\frac{m_{i,t}}{p} g_{i,t}(\theta)\right\|^2\right]
=
\frac{1}{p^2}\,\mathbb{E}_m[m_{i,t}]\,\|g_{i,t}(\theta)\|^2 \nonumber\\
&=
\frac{1}{p}\,\|g_{i,t}(\theta)\|^2.
\label{eq:urs-second-moment}
\end{align*}
Thus, decreasing \(p\) increases gradient norms (and variance) roughly as \(1/\sqrt{p}\)
in typical regimes, which can trigger more frequent gradient clipping and slow
optimization. Empirically, we find \(p\) can be reduced until clipping becomes
dominant; beyond that point training becomes unstable.

\paragraph{Limitation: URS saves backward FLOPs but not forward compute.}
Crucially, independent token masking does not necessarily reduce forward-pass cost
in standard causal Transformers. Even if token \(t\) is masked out of the loss,
computing \(\log\pi_\theta(o_{i,t}\mid q,o_{i,<t})\) still requires processing all
preceding tokens \(1,\dots,t-1\) due to causal self-attention. As a result, URS
primarily reduces \emph{backward} computation (and some gradient-related memory),
but retains near full \emph{forward} cost. This motivates a structured masking
scheme that is compatible with prefix-truncated computation.

%% file: 4_RPC.tex
\section{Random Prefix Cutting for Efficient Long-Context Training}
\label{sec:random-prefix-cutting}

\paragraph{Motivation.}
Long-CoT training is expensive not only because of backpropagation, but also due
to the forward pass through long contexts. In causal Transformers, forward compute
and activation memory scale with sequence length (and attention scales
quadratically with length under dense attention). To reduce both \emph{forward}
and \emph{backward} costs, we propose a structured sampler that preserves the
prefix dependency structure.

\paragraph{Random Prefix Cutting (RPC)}
Instead of selecting tokens independently, RPC samples a \emph{contiguous prefix}
of each response. For trajectory \(i\) of length \(T_i\), we sample a cutoff
\(L_i \in \{1,\dots,T_i\}\) from a distribution \(q_i(\ell)=\Pr(L_i=\ell)\), and
define the mask
\begin{equation}
m_{i,t} \;=\; \mathbb{I}\{t \le L_i\}.
\label{eq:rpc-mask}
\end{equation}
The induced inclusion probability (survival function) is
\begin{equation}
p_{i,t}
\;=\;
\Pr(L_i \ge t)
\;=\;
\sum_{\ell=t}^{T_i} q_i(\ell),
\qquad t=1,\dots,T_i,
\label{eq:rpc-survival}
\end{equation}
with \(p_{i,1}=1\), \(p_{i,t+1}\le p_{i,t}\), and \(p_{i,T_i}=q_i(T_i)>0\).

\paragraph{HT-corrected unbiased objective.}
RPC uses the same HT form as Eq.~\eqref{eq:ht-estimator}, but with the structured
mask in Eq.~\eqref{eq:rpc-mask}:
\begin{equation}
\widehat{\mu}_i^{\textsc{rpc}}(\theta)
\;=\;
\frac{1}{T_i}\sum_{t=1}^{T_i}\frac{m_{i,t}}{p_{i,t}}\,L^{\text{GRPO}}_{i,t}(\theta)
\;=\;
\frac{1}{T_i}\sum_{t=1}^{L_i}\frac{1}{p_{i,t}}\,L^{\text{GRPO}}_{i,t}(\theta).
\label{eq:rpc-ht}
\end{equation}
By Proposition~\ref{prop:ht-unbiased} and \(\mathbb{E}[m_{i,t}]=p_{i,t}\),
\(\widehat{\mu}_i^{\textsc{rpc}}(\theta)\) is an unbiased estimator of the
full-sequence mean loss, and yields an unbiased estimator of the full-token GRPO
gradient.

\paragraph{Why RPC is fundamentally different from deterministic truncation.}
A deterministic prefix truncation that always keeps the first
\(K_i=\lfloor \beta T_i\rfloor\) tokens corresponds to \(p_{i,t}=1\) for
\(t\le K_i\) and \(p_{i,t}=0\) for \(t>K_i\), which violates the HT requirement
\(p_{i,t}>0\) and induces systematic bias by permanently discarding late-token
contributions. In reasoning tasks, late tokens often contain verification steps
or final-answer formation; ignoring them can distort the learned policy and
degrade convergence. RPC avoids this failure mode by ensuring every position has
nonzero inclusion probability while still enabling prefix-truncated computation.

\paragraph{Compute savings: prefix structure enables true forward truncation.}
Unlike URS, RPC can reduce forward compute because the model only needs to process
the prefix of length \(L_i\). Under dense attention, the per-sequence attention
cost reduces from \(O(T_i^2)\) to \(O(L_i^2)\), and activation memory scales with
\(L_i\) rather than \(T_i\). Thus RPC provides savings in \emph{both} the forward
and backward passes, while remaining statistically unbiased via HT correction.

\paragraph{Mask dependence and variance.}
RPC masks are positively correlated: if \(m_{i,t}=1\) then all earlier positions
are also included. For \(s \le t\),
\begin{equation*}
\label{eq:rpc-cov}
\begin{aligned}
\mathrm{Cov}(m_{i,s},m_{i,t})
&=
\Pr(L_i \ge t)-\Pr(L_i \ge s)\Pr(L_i \ge t) \\
&=
p_{i,t}\,(1-p_{i,s})
\;\ge\; 0.
\end{aligned}
\end{equation*}
This dependence can increase variance relative to idealized independent sampling,
but it is precisely what makes RPC compatible with prefix-truncated computation.
Practically, stability is controlled by designing \(q_i(\ell)\) so that
\(p_{i,t}\) does not decay too sharply, avoiding extreme importance weights
\(1/p_{i,t}\) near the end of the sequence. This is also to ensure that the model will get a fair amount of exposure to later parts of the trajectory. 

\paragraph{Uniform cutoff as a simple default.}
A robust choice is the uniform cutoff \(L_i \sim \mathrm{Uniform}(\{1,\dots,T_i\})\),
i.e., \(q_i(\ell)=1/T_i\). Then
\begin{equation*}
p_{i,t}=\frac{T_i-t+1}{T_i},
\qquad
\mathbb{E}[L_i]=\sum_{t=1}^{T_i}p_{i,t}=\frac{T_i+1}{2}\approx \frac{T_i}{2}.
\label{eq:rpc-uniform}
\end{equation*}
On average, RPC processes about half the tokens per trajectory, leading to
substantial reductions in forward/backward cost while preserving unbiased access
to gradients from all positions through randomization. The variance analysis of Uniform cutoff is further discussed in Appendix \ref{app:rpc-details}.

\paragraph{Minimum-cutoff RPC}
In some implementations we enforce a minimum retained prefix length \(C\), to avoid extremely short prefixes and to stabilize the HT weights near the beginning of training. Concretely, we sample
\(
L_i \sim \mathrm{Uniform}(\{C,\dots,T_i\})
\)
(or more generally any distribution supported on \([C,T_i]\)), and keep
\(m_{i,t}=\mathbb{I}\{t\le L_i\}\).
Under the uniform choice, the survival probabilities become
\[
p_{i,t}=\Pr(L_i\ge t)=
\begin{cases}
1, & t\le C,\\[2pt]
\frac{T_i-t+1}{T_i-C+1}, & t>C,
\end{cases}
\]
which guarantees bounded importance weights \(1/p_{i,t}\le \frac{T_i-C+1}{T_i-t+1}\) for all included positions and strictly prevents pathological ``one-token'' prefixes. The expected retained length is
\[
\mathbb{E}[L_i]
=
\frac{C+T_i}{2}
\;=\;
\frac{T_i}{2}+\frac{C}{2},
\]
so compared to the unconstrained uniform cutoff (which has \(\mathbb{E}[L_i]\approx T_i/2\)), enforcing a minimum cutoff increases the expected compute by an additive \(C/2\) tokens per sequence while often improving stability.

\begin{table*}[t]
\caption{Comparison of token-efficient methods. RPC is the only method that provides computational savings in both passes while remaining statistically unbiased.}
\centering
\small
\begin{tabular}{lcccl}
\hline
\textbf{Method} & \textbf{Unbiased?} & \textbf{Forward Savings} & \textbf{Backward Savings} & \textbf{Key Property} \\
\hline
URS & Yes & No & Yes & Simple, constant $p$ sampling \\
Det. Trunc. & No & Yes & Yes & Systematic bias, ignores late tokens \\
\textbf{RPC} & \textbf{Yes} & \textbf{Yes} & \textbf{Yes} & Structured, preserves causal context \\
\hline
\end{tabular}
\label{tab:method-comparison}
\end{table*}

\paragraph{Summary.}
Token masking with HT reweighting yields unbiased gradient estimates while trading
compute for variance. URS provides primarily backward savings but can inflate
gradient norms as \(p\) decreases. RPC preserves prefixes, enabling \emph{true}
forward truncation and thus savings in both passes, while remaining unbiased and
avoiding the systematic bias of deterministic truncation.

%% file: 5_Exps.tex
\section{Experiments and Results}

\subsection{Experiment Setup}
In this section, we adopt \textbf{Verl} \cite{sheng2025hybridflow} as our training framework. The training dataset is DAPO-Math-17K from DAPO paper \cite{yu2025dapo}.
The models we consider are \textbf{Qwen2.5-Math-7B} model \cite{yang2024qwen2} and \textbf{Qwen3-8B} model \cite{yang2025qwen3}. The RL algorithm we adopt is GRPO \cite{shao2024deepseekmath} without KL Divergence as suggested by DAPO \cite{yu2025dapo}.  The corresponding training recipes are provided in Appendix \ref{app: train_recipe}. All methods are trained with the same number of optimizer updates and identical optimization hyperparameters.

\paragraph{Algorithms}
The algorithms we used in this experiment are: 
\begin{itemize}
    \item \textbf{GRPO}: vanilla GRPO algorithm \cite{shao2024deepseekmath} using full response tokens. 
    \item \textbf{URS}: GRPO with random token masking using uniform random sampling with $p=0.5$.
    \item \textbf{Det. Trunc.}: GRPO with deterministic prefix truncation of 50\% of trajectory tokens. 
    \item \textbf{RPC}: GRPO with uniform random prefix cutting with a minimum of 100 tokens.  
\end{itemize}
Each algorithm is repeated \textbf{5} times to capture the variance of validation metrics. Given the limited GPU resources, we randomly sample 80\% of DAPO-Math-17K dataset once and reuse it for all runs to speed up training. The training epoch is selected when the reward curve of \textbf{GRPO} is plateauing. In our case, we choose 10 epoch for Qwen2.5-Math-7B and 2 epoch for Qwen3-8B. All hyperparameters are shared across algorithms to ensure a fair comparison.

\paragraph{Evaluation}We evaluate our algorithms on standard mathematical reasoning benchmarks: \textbf{AIME24, AIME25, MATH500} \cite{hendrycks2021measuring}. For each question, we generate 16 independent responses under a decoding temperature T = 1.0, and
report the average accuracy and the average number of tokens per response.

\begin{table*}[t]
  \caption{Token-efficient RL training results on three math benchmarks (MATH, AIME24, AIME25). We report Acc@16 and pass@16 (mean $\pm$ 95\% confidence interval across runs) for two base models. Within each model, we color each cell by whether its 95\% CI overlaps the GRPO baseline for the same metric (\cig{green}: overlap; \cib{red}: non-overlap, \cir{grey}: overlap with much high variance and lower scores).}
  \label{tab:acc16_pass16_two_models}
  \centering
  \small
  \setlength{\tabcolsep}{4.5pt}
  \renewcommand{\arraystretch}{1.10}
  \begin{tabular}{l|cc|cc|cc}
    \toprule
    \textbf{Method} & \multicolumn{2}{c|}{\textbf{MATH}} & \multicolumn{2}{c|}{\textbf{AIME24}} & \multicolumn{2}{c}{\textbf{AIME25}} \\
     & Acc@16 & pass@16 & Acc@16 & pass@16 & Acc@16 & pass@16 \\
    \midrule

    \rowcolor{modelbgA}
    \multicolumn{7}{c}{\textbf{Qwen2.5-Math-7B Model}} \\
    \midrule
    \textbf{GRPO} & 0.610$\pm$0.031 & 0.710$\pm$0.064 & 0.259$\pm$0.027 & 0.469$\pm$0.037 & 0.126$\pm$0.020 & 0.256$\pm$0.015 \\
    URS &
      \cig{0.610$\pm$0.050} & \cig{0.741$\pm$0.080} &
      \cig{0.190$\pm$0.066} & \cig{0.360$\pm$0.105} &
      \cig{0.116$\pm$0.051} & \cig{0.229$\pm$0.062} \\
    Det.\ Trunc. &
      \cir{0.361$\pm$0.265} & \cir{0.423$\pm$0.318} &
      \cib{0.037$\pm$0.076} & \cib{0.072$\pm$0.159} &
      \cib{0.027$\pm$0.071} & \cir{0.071$\pm$0.178} \\
    RPC &
      \cig{0.670$\pm$0.093} & \cig{0.807$\pm$0.135} &
      \cig{0.170$\pm$0.079} & \cig{0.405$\pm$0.081} &
      \cig{0.122$\pm$0.019} & \cig{0.282$\pm$0.065} \\
    \midrule

    \rowcolor{modelbgA}
    \multicolumn{7}{c}{\textbf{Qwen3-8B Model}} \\
    \midrule
    \textbf{GRPO} & 0.768$\pm$0.003 & 0.911$\pm$0.005 & 0.257$\pm$0.017 & 0.475$\pm$0.024 & 0.202$\pm$0.011 & 0.386$\pm$0.040 \\
    URS &
      \cig{0.768$\pm$0.006} & \cig{0.911$\pm$0.009} &
      \cig{0.245$\pm$0.009} & \cig{0.474$\pm$0.056} &
      \cig{0.207$\pm$0.010} & \cig{0.384$\pm$0.044} \\
    Det.\ Trunc. &
      \cib{0.633$\pm$0.004} & \cig{0.902$\pm$0.004} &
      \cib{0.190$\pm$0.013} & \cig{0.488$\pm$0.033} &
      \cib{0.159$\pm$0.014} & \cig{0.374$\pm$0.029} \\
    RPC &
      \cig{0.764$\pm$0.010} & \cig{0.910$\pm$0.009} &
      \cig{0.236$\pm$0.014} & \cig{0.442$\pm$0.040} &
      \cig{0.201$\pm$0.018} & \cig{0.370$\pm$0.030} \\
    \bottomrule
  \end{tabular}
\end{table*}

\begin{figure}[t]
  \centering
  \begin{subfigure}[t]{0.49\columnwidth}
    \centering
    \includegraphics[width=\linewidth]{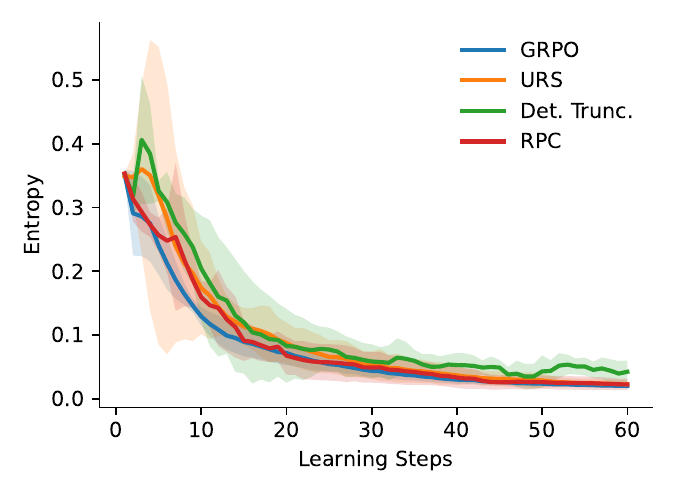}
    \caption{Qwen2.5-Math-7B}
    \label{fig:qwen_2_5_entropy}
  \end{subfigure}
  \hfill
  \begin{subfigure}[t]{0.49\columnwidth}
    \centering
    \includegraphics[width=\linewidth]{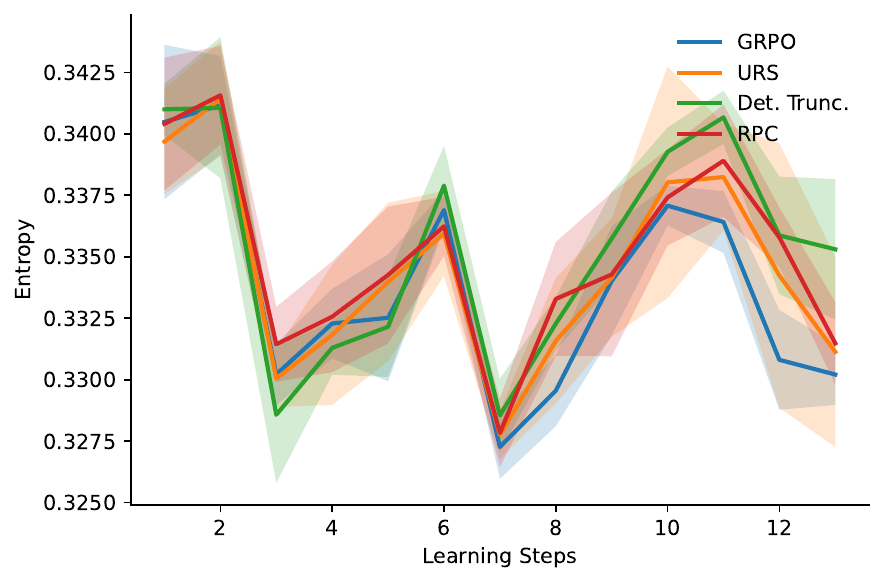}
    \caption{Qwen3-8B}
    \label{fig:qwen_3_entropy}
  \end{subfigure}
 \caption{
      Entropy curves with 95\% confidence interval across 5 runs for \textbf{GRPO} (vanilla GRPO), \textbf{URS} (GRPO with random sampling $p=0.5$), \textbf{Det. Trunc.} (GRPO with deterministic prefix truncation of 50\% of trajectory tokens) and \textbf{RPC} (GRPO with uniform random prefix cutting). }
  \label{fig:entropy}
\end{figure}

\begin{figure}[t]
  \centering
  \begin{subfigure}[t]{0.49\columnwidth}
    \centering
    \includegraphics[width=\linewidth]{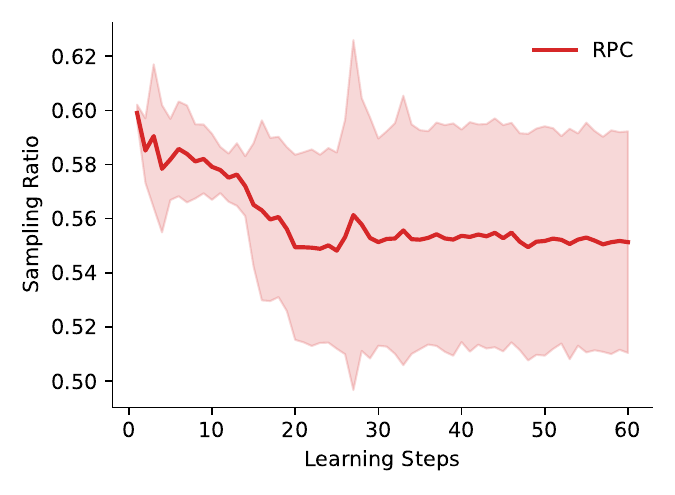}
    \caption{Qwen2.5-Math-7B}
    \label{fig:qwen_2_5_ratio}
  \end{subfigure}
  \hfill
  \begin{subfigure}[t]{0.49\columnwidth}
    \centering
    \includegraphics[width=\linewidth]{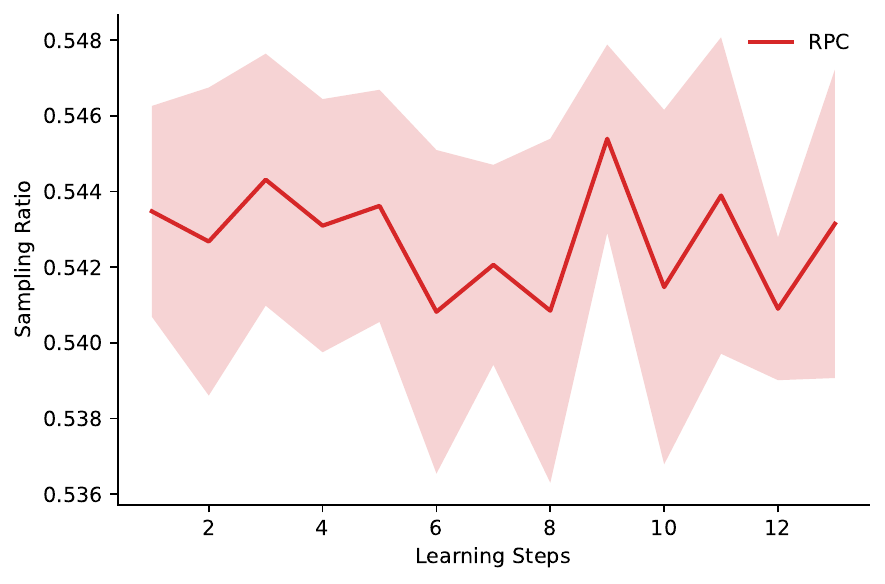}
    \caption{Qwen3-8B}
    \label{fig:qwen_3_ratio}
  \end{subfigure}
 \caption{
      Percentage of selected tokens with 95\% confidence interval across 5 runs for \textbf{RPC} (GRPO with uniform random prefix cutting). }
  \label{fig:ratio}
\end{figure}

\begin{table*}[t]
  \caption{System efficiency metrics for token-efficient RL training (mean $\pm$ 95\% CI across 5 runs). 
  We report average peak GPU memory, average training time per step excluding inference, and average total time per step including inference, each averaged over training steps. 
  Within each model block, URS overlaps with GRPO on all three metrics under 95\% CIs. 
  \cig{Green} indicates better efficiency (lower mean) than GRPO; \cir{gray} indicates overlap/parity with GRPO.}
  \label{tab:efficiency_two_models_corrected}
  \centering
  \small
  \setlength{\tabcolsep}{4.5pt}
  \renewcommand{\arraystretch}{1}
  \begin{tabular}{l|c|c|c}
    \toprule
  \textbf{Method}
& \makecell[c]{\textbf{Avg. Peak GPU}\\\textbf{Memory (GB)}}
& \makecell[c]{\textbf{Avg. Train Time / Step (s)}\\\textbf{w/o inference}}
& \makecell[c]{\textbf{Avg. Total Time}\\\textbf{/ Step (s)}} \\
    \midrule

    \rowcolor{modelbgA}
    \multicolumn{4}{c}{\textbf{Qwen2.5-Math-7B Model}} \\
    \midrule
    \textbf{GRPO}
      & 35.810$\pm$0.042
      & 166.579$\pm$6.728
      & 291.240$\pm$11.943 \\
    URS
      & \cig{35.758$\pm$0.136}
      & \cig{147.399$\pm$34.520}
      & \cig{259.376$\pm$55.345} \\
    Det.\ Trunc.
      & \cig{24.603$\pm$0.487}
      & \cig{81.552$\pm$4.322}
      & \cig{137.211$\pm$9.388} \\
    RPC
      & \cig{29.235$\pm$0.362}
      & \cig{121.918$\pm$9.601}
      & \cig{230.428$\pm$17.573} \\
    \midrule

    \rowcolor{modelbgA}
    \multicolumn{4}{c}{\textbf{Qwen3-8B Model}} \\
    \midrule
    \textbf{GRPO}
      & 47.718$\pm$0.215
      & 311.502$\pm$2.555
      & 628.027$\pm$7.008 \\
    URS
      & \cir{47.802$\pm$0.144}
      & \cir{312.439$\pm$1.525}
      & \cir{631.863$\pm$2.692} \\
    Det.\ Trunc.
      & \cig{33.786$\pm$0.009}
      & \cig{183.114$\pm$0.304}
      & \cig{364.594$\pm$0.934} \\
    RPC
      & \cig{39.234$\pm$0.852}
      & \cig{220.322$\pm$2.120}
      & \cig{400.547$\pm$4.201} \\
    \bottomrule
  \end{tabular}
\end{table*}

\paragraph{Token-Efficient RL on Accuracy.}
Table~\ref{tab:acc16_pass16_two_models} shows that \textsc{Det.\ Truncation} is directionally worse than full-token \textsc{GRPO} on most metrics, often by a large margin in point estimates. 
Although several gaps are not statistically significant under the 95\% CI overlap heuristic, this is primarily because \textsc{Det.\ Truncation} exhibits much higher run-to-run variance for Qwen2.5-Math-7B Model. For Qwen3-8B model, all Acc@16 metrics are significantly lower than \textsc{GRPO} while Pass@16 metrics are not. This is because the base model is strong and reinforcement learning algorithm does not lift pass@k at large k as discussed in \citep{yue2025does}.
This behavior is consistent with the expected bias of deterministic truncation: always removing the same suffix can suppress late-step learning signals and destabilize optimization. 
By contrast, the unbiased sampling-based methods, \textsc{URS} and \textsc{RPC}, remain consistently on par with \textsc{GRPO} across MATH, AIME24, and AIME25 for both models, with overlapping 95\% CIs throughout.

\paragraph{Token-Efficient RL on Entropy.}
From entropy Figure \ref{fig:entropy}, \textsc{Det.\ Truncation} exhibits consistently higher entropy, likely due to biased gradients and reduced effective training signal. For \textsc{URS} and \textsc{RPC}, entropy curves converge to same values as \textsc{GRPO}.

\paragraph{Token-Efficient RL on GPU memory.}
 Although absolute memory usage can vary with hardware and training configuration (e.g., batch size and maximum response length), all methods are evaluated under the same setup (Appendix~\ref{app: train_recipe}), enabling a fair comparison. To quantify memory efficiency during RL training, we use VERL’s runtime metric \texttt{allocated\_memory\_gb} \cite{sheng2025hybridflow}, which records the peak GPU memory allocated at each training step. Table~\ref{tab:efficiency_two_models_corrected} show a clear pattern. \textsc{Det.\ Truncation} is the most memory-efficient method because it deterministically keeps only the first 50\% of tokens, thereby avoiding activation materialization for the truncated suffix. However, this memory advantage comes with a clear quality cost: as shown in Table~\ref{tab:acc16_pass16_two_models}, \textsc{Det.\ Truncation} consistently underperforms \textsc{GRPO} on accuracy metrics, with high variance in several settings. By contrast, \textsc{URS} (uniform random token sampling) yields little to no peak-memory reduction relative to \textsc{GRPO}. This is expected: \textsc{URS} reduces the number of tokens contributing to the loss/backward signal, but does not shorten the effective forward-prefix computation induced by causal attention, so peak activation memory remains close to full-token training. \textsc{RPC} achieves the best trade-off. By randomly cutting contiguous prefixes and applying Horvitz--Thompson correction, it preserves statistical alignment with full-token optimization while reducing the effective sequence length used in training computation. Empirically, \textsc{RPC} remains on par with \textsc{GRPO} on accuracy across benchmarks, while reducing peak GPU memory by approximately 18--20\% in our runs (e.g., 35.81$\rightarrow$29.24~GB on Qwen2.5-Math-7B and 47.72$\rightarrow$39.23~GB on Qwen3-8B). Overall, these results indicate that \textsc{RPC} provides the most practical memory--quality trade-off among the compared methods.

\paragraph{Token-Efficient RL on Training Time.}
From Table~\ref{tab:efficiency_two_models_corrected}, for \textbf{Qwen2.5-Math-7B}, both sampling-based methods are faster than full-token \textsc{GRPO}. Relative to \textsc{GRPO}, \textsc{URS} reduces update time (w/o inference) by about \textbf{11.5\%} and total step time by about \textbf{10.9\%}. \textsc{RPC} delivers larger gains, reducing update time by about \textbf{26.8\%} and total step time by about \textbf{20.9\%}.  For \textbf{Qwen3-8B}, \textsc{URS} is effectively on par with \textsc{GRPO} in timing (differences are within roughly \textbf{1\%} and 95\% CIs overlap), whereas \textsc{RPC} remains substantially faster: update time decreases by about \textbf{29.3\%} and total step time by about \textbf{36.2\%}. Overall, these results align with NAT’s design objective: token-efficient optimization mainly accelerates the learner path while leaving rollout inference unchanged. As a result, improvements are most pronounced in the no-inference metric, and convert into clear end-to-end gains when learner-side cost constitutes a meaningful portion of each step. Across both model scales, \textsc{RPC} provides the largest and most consistent wall-clock improvements while maintaining competitive accuracy (Table~\ref{tab:acc16_pass16_two_models}).

\paragraph{Selected-token ratio under RPC}
Figure~\ref{fig:ratio} shows that RPC retains approximately half of each trajectory on average, with the selected-token ratio stabilizing around \(\sim 0.54\!-\!0.56\) across training. This is slightly above \(50\%\) by design: our implementation uses uniform random prefix cutting with a minimum retained prefix length \(C=100\),
so the expected selection ratio is strictly larger than \(0.5\), with the additive term \(C/(2T_i)\) most visible for shorter responses. Empirically, the observed \(\sim 54\%-56\%\) range is consistent with this prediction and confirms that RPC operates in the intended \(\approx 50\%\) token-budget regime while avoiding pathological ultra-short prefixes.


%% file: 6_Related_Work.tex
\section{Related Work}

\paragraph{Efficient RL Training.} Recent profiling of GRPO training reveals that the rollout generation phase accounts for approximately \textbf{70\%} of total training time \citep{gao2025rollpacker, zhou2025april}. This bottleneck is primarily driven by ``long-tail rollouts,'' where high variance in sequence lengths leads to significant GPU under-utilization and pipeline bubbles. Systems such as \textsc{RollPacker} \citep{gao2025rollpacker} and \textsc{SortedRL} \citep{zhang2025sortedrl} mitigate this via length-aware scheduling, isolating straggling sequences to maximize throughput. \textsc{April} \citep{zhou2025april} further reduces latency through over-provisioning rollout requests and terminating stragglers once a target sample quota is met. To accelerate the generation itself, \textsc{Tlt} \citep{hu2025taming} leverages idle GPU cycles to train on-the-fly drafter models for speculative decoding. \citet{xu2025not} proposed the down-sampling rollouts and focusing on subset of trajectories which can maintain performance while significantly reducing gradient computation.  Our work is orthogonal to these rollout optimizations; by focusing on token-level efficiency during the forward and backward passes, our method can be seamlessly integrated with existing rollout acceleration frameworks.

 \paragraph{Partial Tokens Training.} Beyond system-level scheduling, recent research has explored selective token-level updates to reduce the computational cost of the forward and backward passes. \citet{wang2025beyond} demonstrate that RL performance is driven by a ``high-entropy minority'' of tokens, suggesting that gradients from low-entropy tokens are largely redundant for reasoning tasks. Most relevantly, \citet{lee2025token} propose token-efficient RL schemes that mask random tokens. Our work extends this line of research by introducing a framework using reweighting to offer a unified theory analysis. Moreover, random masking and entropy-based masking cannot reduce forward pass cost since activation memory and forward computation are still done using full tokens. Our approach ensures that we match the full-sequence statistics while strictly capping the maximum sequence length processed by the policy network, leading to predictable and substantial memory savings.






%% file: 7_Limitation.tex
\section{Limitation and Future Work}
Beyond the empirical gains, there are two conceptual reasons why \textsc{RPC} is effective.

First, \textsc{RPC} can be interpreted as a \emph{structured regularizer} analogous in spirit to dropout: instead of randomly dropping individual hidden units, it randomly drops suffix segments while preserving causal prefix structure. This introduces stochasticity in the optimization path and reduces over-reliance on any single long trajectory realization, while remaining compatible with autoregressive decoding constraints. In that sense, \textsc{RPC} acts like sequence-level, causality-preserving dropout \citep{srivastava2014dropout} rather than token-independent masking. 

Second, \textsc{RPC} is aligned with long-horizon credit-assignment dynamics. In many reasoning trajectories, early decisions constrain downstream token distributions; thus, late-token gradients can be high-cost but lower-marginal-value or noisier for policy improvement. Randomized prefix training increases exposure to early causal decisions across updates, while avoiding the systematic bias of deterministic fixed-length truncation. This offers a practical bias--efficiency balance for long-CoT RL, where temporal credit assignment is known to be a central challenge \cite{pignatelli2023survey}. 

That said, our current study only instantiates two simple samplers (\textsc{URS} and \textsc{RPC}). Both are \emph{information-agnostic}: they do not use token-level uncertainty, gradient magnitude, or verifier-derived saliency. A key future direction is to learn or adapt inclusion probabilities within the same Horvitz--Thompson framework so that compute is preferentially allocated to high-information tokens, potentially reducing estimator variance at fixed token budget. Another direction is to co-design sampling with systems kernels (e.g., blockwise/prefix-aware attention and memory scheduling) to further improve end-to-end efficiency. Finally, while our results are strong on math RLVR, broader validation is still needed on domains with different reward sparsity, trajectory lengths, and verifier reliability. These observations motivate a broader research agenda on information-aware token selection and systems co-design. We now summarize the main takeaways of this work.

%% file: 8_Discussion.tex
\section{Conclusion}
\label{sec:conclusion}

In this work, we challenged the conventional assumption that full-sequence backpropagation is necessary for effective reinforcement learning in long-form reasoning tasks. We introduced \textsc{Nat} (Not All Tokens are Needed), a unified framework that utilizes Horvitz-Thompson estimation to decouple sequence-level reward evaluation from token-level policy optimization. By leveraging unbiased gradient estimators, \textsc{Nat} enables selective updates on trajectory subsets—such as high-entropy clusters or randomized prefixes—without inducing the systematic bias inherent in naive deterministic truncation. Our empirical results across challenging mathematical benchmarks demonstrate that \textsc{Nat} achieves performance parity with full-token GRPO while utilizing as few as \textbf{50\%} of the trajectory tokens. In particular, the Random Prefix Cutting (RPC) strategy provides a scalable solution to the memory and compute bottlenecks of long chain-of-thought training by significantly reducing activation memory and increasing training throughput. As reasoning trajectories continue to grow in complexity, \textsc{Nat} offers a theoretically grounded and orthogonal approach to existing rollout optimizations, providing a vital tool for the efficient scaling of frontier AI systems.

%% file: 9_Appendix.tex
\section{Gradient Unbiasedness}
\label{app:gradient-unbias}

\begin{proof}[Proof of Gradient Unbiasedness]
Given the Horvitz--Thompson estimator $\widehat{\mu}_i^{\text{HT}}(\theta)$, its gradient is:
\begin{equation*}
\nabla_\theta \widehat{\mu}_i^{\text{HT}}(\theta) = \frac{1}{T_i}\sum_{t=1}^{T_i} \frac{m_{i,t}}{p_{i,t}} \nabla_\theta L_{i,t}^{\text{GRPO}}(\theta).
\end{equation*}
Taking expectation over the mask distribution:
\begin{align*}
\mathbb{E}_m\left[\nabla_\theta \widehat{\mu}_i^{\text{HT}}(\theta)\right] 
&= \frac{1}{T_i}\sum_{t=1}^{T_i} \frac{\mathbb{E}[m_{i,t}]}{p_{i,t}} \nabla_\theta L_{i,t}^{\text{GRPO}}(\theta) \\
&= \frac{1}{T_i}\sum_{t=1}^{T_i} \nabla_\theta L_{i,t}^{\text{GRPO}}(\theta) \\
&= \nabla_\theta \mu_i(\theta).
\end{align*}
Thus, the gradient estimator is unbiased for the full-token gradient.
\end{proof}

\section{Appendix: Detailed Derivations and Extensions for Random Prefix Cutting}
\label{app:rpc-details}

\subsection{Variance Derivation}
\label{app:rpc-variance-derivation}

We analyze the variance of the RPC Horvitz--Thompson estimator by explicitly accounting for the
\emph{stopping-time structure} induced by prefix truncation.
Let
\[
\widehat{\mu}_i^{\text{RPC-HT}}
=
\frac{1}{T_i}
\sum_{t=1}^{T_i}
\frac{m_{i,t}}{p_{i,t}}
L_{i,t}^{\text{GRPO}}(\theta),
\]
where the binary mask $m_{i,t}$ indicates whether the rollout survives up to token $t$.

\paragraph{Stopping-time formulation.}
Define the random stopping time
\[
\tau_i \;\triangleq\; \max\{t : m_{i,t} = 1\},
\]
with the convention that $\tau_i = 0$ if the rollout terminates immediately.
By construction,
\[
m_{i,t} = \mathbf{1}\{t \le \tau_i\},
\qquad
\mathbb{P}(\tau_i \ge t) = p_{i,t},
\]
and the estimator can be rewritten as
\begin{equation}
\widehat{\mu}_i^{\text{RPC-HT}}
=
\frac{1}{T_i}
\sum_{t=1}^{\tau_i}
\frac{L_{i,t}^{\text{GRPO}}(\theta)}{p_{i,t}}.
\label{eq:rpc-ht-stopping}
\end{equation}

\paragraph{Variance expression.}
Using~\eqref{eq:rpc-ht-stopping}, the variance admits the exact decomposition
\begin{equation}
\mathrm{Var}\!\left[\widehat{\mu}_i^{\text{RPC-HT}}\right]
=
\frac{1}{T_i^2}
\left(
\mathbb{E}
\left[
\left(
\sum_{t=1}^{\tau_i}
\frac{L_{i,t}^{\text{GRPO}}(\theta)}{p_{i,t}}
\right)^2
\right]
-
\left(
\sum_{t=1}^{T_i}
L_{i,t}^{\text{GRPO}}(\theta)
\right)^2
\right).
\label{eq:rpc-variance-stopping}
\end{equation}
This form is exact and does not rely on token-level independence assumptions.

\paragraph{Discussion.}
Unlike standard Horvitz--Thompson estimators with fixed sampling designs,
the random variables $\{m_{i,t}\}_{t=1}^{T_i}$ are deterministically coupled through the stopping
time $\tau_i$: an early termination event sets all subsequent masks to zero.
As a consequence, the variance of the RPC estimator cannot be decomposed into a sum of
independent or pairwise token-level contributions.
Instead, variance is dominated by early stopping events and scales inversely with the
smallest survival probabilities $p_{i,t}$, particularly for late tokens.

This stopping-time coupling explains the elevated variance observed when early prefixes
are dropped with high probability, and motivates conservative scheduling of survival
probabilities in practice.

\subsection{Optional Mandatory Prefix}
\label{app:mandatory-prefix}

In practice, to reduce variance and guarantee that initial reasoning steps are always observed, we can enforce that the first $C$ tokens are always included. This is done by restricting $L_i \in \{C, C+1, \dots, T_i\}$ and setting $p_{i,t}=1$ for $t \le C$. The survival function for $t > C$ remains $p_{i,t} = \Pr(L_i \ge t)$. The unbiased estimator~\eqref{eq:ht-estimator} still applies. For a uniform cutoff over $\{C,\dots,T_i\}$, we have
\begin{equation}
\mathbb{E}[L_i] = \frac{C + T_i}{2}, \qquad
\frac{\mathbb{E}[L_i]}{T_i} = \frac{1}{2} + \frac{C}{2T_i}.
\label{eq:uniform-cutoff-with-C}
\end{equation}

\subsection{Optimal Survival Schedule under a Budget}
\label{app:optimal-schedule}

Given a compute budget $B = \sum_{t=1}^{T_i} p_{i,t}$, we might seek the survival schedule $\{p_{i,t}\}$ that minimizes the variance of the estimator. Ignoring the off-diagonal correlation terms (or assuming token losses are uncorrelated), the variance is proportional to $\sum_{t=1}^{T_i} \frac{\tilde{L}_{i,t}^2}{p_{i,t}}$. Using Lagrange multipliers to minimize this sum subject to $\sum_t p_{i,t} = B$ and monotonicity ($1 = p_{i,1} \ge \dots \ge p_{i,T_i} > 0$) yields the intuitive solution $p_{i,t}^* \propto |\tilde{L}_{i,t}|$. However, the losses $\tilde{L}_{i,t}$ are unknown before the forward pass. Without prior knowledge, the uniform cutoff (linear $p_{i,t}$) emerges as a robust max-entropy choice, spreading probability mass evenly across feasible prefix lengths and minimizing worst-case variance.

\subsection{Comparison with Independent Token Masking}
\label{app:independent-masking}

For an independent masking scheme where each token $t$ is kept independently with probability $p_{i,t}$, the covariance is $\mathrm{Cov}(m_{i,s}, m_{i,t}) = 0$ for $s \neq t$, and $\mathrm{Var}(m_{i,t}) = p_{i,t}(1-p_{i,t})$. The variance of the Horvitz--Thompson estimator then becomes
\begin{equation}
\mathrm{Var}_{\text{indep}} = \frac{1}{T_i^2} \sum_{t=1}^{T_i} \tilde{L}_{i,t}^2 \frac{1-p_{i,t}}{p_{i,t}}.
\label{eq:rpc-variance-explicit-app}
\end{equation}
Comparing with Eq.~\eqref{eq:rpc-variance-explicit-app}, the RPC variance includes an extra positive term due to the covariances. Nevertheless, as argued in the main text, independent masking often leads to higher \emph{effective} variance in practice because of unstable importance weights and the necessity to run the full forward pass (no computational saving).

\subsection{Bias-Variance Trade-off Formalization}
\label{app:bias-variance-formal}

The mean squared error (MSE) of a gradient estimator $\hat{\mu}$ relative to the true full-sequence gradient $\mu$ is
\[
\mathrm{MSE}(\hat{\mu}) = \mathbb{E}\|\hat{\mu} - \mu\|^2 = \mathrm{Var}(\hat{\mu}) + \|\mathbb{E}[\hat{\mu}] - \mu\|^2.
\]
Deterministic truncation has low variance but high bias squared $\|\mathbb{E}[\hat{\mu}_{\text{det}}] - \mu\|^2$, which does not vanish with more steps. RPC has zero bias ($\mathbb{E}[\hat{\mu}_{\text{RPC}}] = \mu$) but higher variance. Stochastic optimization algorithms (e.g., SGD) are robust to zero-mean noise (variance) but can diverge or converge to wrong solutions under persistent bias. Hence, the unbiasedness of RPC is critical for correct convergence.

\section{Training Recipe}
\label{app: train_recipe}

\subsection{Training Setup}

We employ Group Relative Policy Optimization (GRPO) to fine-tune mathematical reasoning models using reinforcement learning from outcome-based rewards. Our training infrastructure utilizes 16 H100 GPUs with distributed training via Fully Sharded Data Parallel (FSDP) and asynchronous rollout generation powered by SGLang.

\subsection{Model Configuration}
\begin{itemize}
    \item \textbf{Gradient checkpointing}: Enabled to reduce activation memory at the cost of recomputation during backward passes.
    \item \textbf{Parameter offloading}: Model parameters are offloaded to CPU memory when not actively used.
    \item \textbf{Optimizer offloading}: Optimizer states (Adam moments) are offloaded to CPU to further reduce GPU memory pressure.
    \item \textbf{Fused kernels}: Optimized CUDA kernels for improved throughput.
\end{itemize}

\subsection{Training Hyperparameters}

Our training configuration uses the following key hyperparameters for model \textbf{Qwen 2.5 Math 7B}:
\begin{itemize}
    \item \textbf{Total epochs}: 10
    \item \textbf{Training batch size}: 2048 samples
    \item \textbf{Mini-batch size}: 1024 samples (for advantage computation)
    \item \textbf{Micro-batch size per GPU}: 16 sequences
    \item \textbf{Learning rate}: $1 \times 10^{-5}$ with AdamW optimizer
    \item \textbf{Temperature}: 1.0 for both training and validation sampling
    \item \textbf{Maximum prompt length}: 1024 tokens
    \item \textbf{Maximum response length}: 3000 tokens (total allowed context length is 4096)
    \item \textbf{Rollout samples per prompt}: 8 responses generated per training iteration
    \item \textbf{Validation samples}: 16 responses per prompt for evaluation
    \item \textbf{GPU memory utilization}: 0.75 for rollout generation
    \item \textbf{KL penalty}: Disabled ($\lambda_\text{KL} = 0$) to focus purely on outcome rewards
\end{itemize}

For \textbf{Qwen3-8B} model, we use:
\begin{itemize}
    \item  \textbf{Enable thinking}: False
    \item \textbf{Total epochs}: 2
    \item \textbf{Training batch size}: 1024 samples
    \item \textbf{Mini-batch size}: 1024 samples (for advantage computation)
    \item \textbf{Micro-batch size per GPU}: 16 sequences
    \item \textbf{Learning rate}: $5 \times 10^{-7}$ with AdamW optimizer
    \item \textbf{Temperature}: 1.0 for both training and validation sampling
    \item \textbf{Maximum prompt length}: 1024 tokens
    \item \textbf{Maximum response length}: 8192 tokens
    \item \textbf{Rollout samples per prompt}: 8 responses generated per training iteration
    \item \textbf{Validation samples}: 16 responses per prompt for evaluation
    \item \textbf{GPU memory utilization}: 0.75 for rollout generation
    \item \textbf{KL penalty}: Disabled ($\lambda_\text{KL} = 0$) to focus purely on outcome rewards
\end{itemize}

\section{More Experiment Results}
\label{sec:more_result}

\begin{figure*}
  \centering
  \begin{subfigure}[t]{0.4\columnwidth}
    \centering
    \includegraphics[width=\linewidth]{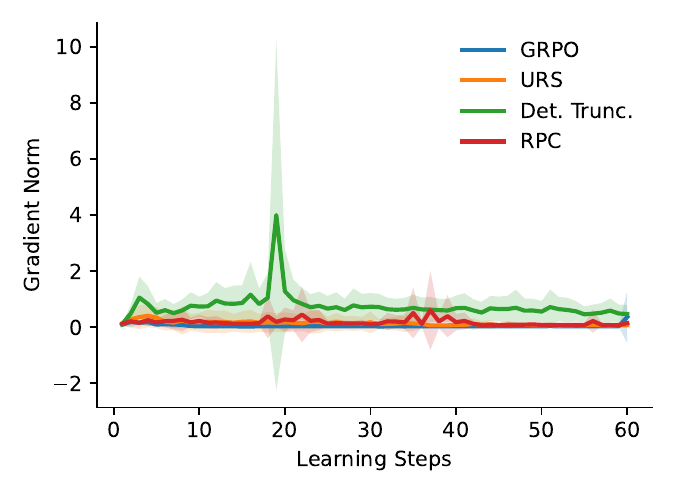}
    \caption{Qwen2.5-Math-7B}
    \label{fig:qwen_2_5_grad_norm}
  \end{subfigure}
  \hfill
  \begin{subfigure}[t]{0.4\columnwidth}
    \centering
    \includegraphics[width=\linewidth]{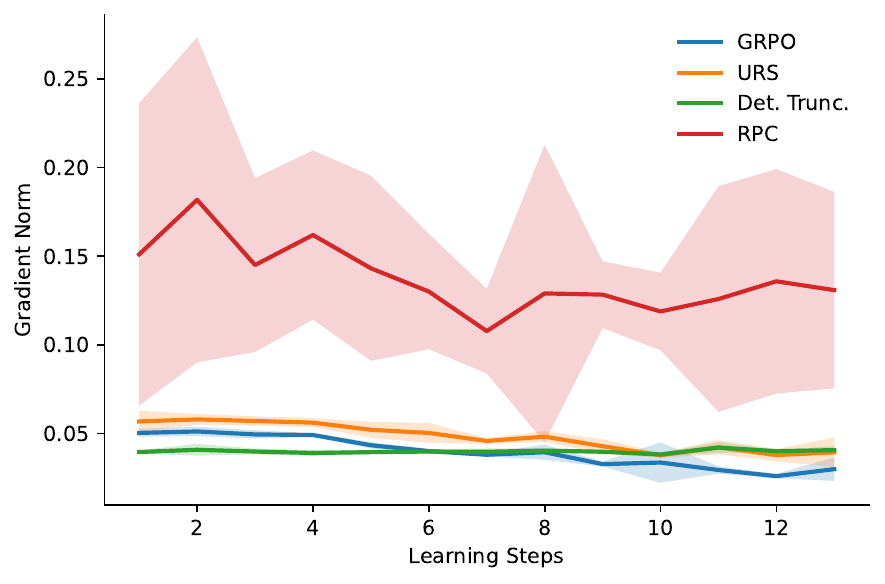}
    \caption{Qwen3-8B}
    \label{fig:qwen_3_grad_norm}
  \end{subfigure}
 \caption{
      Gradient norm curves with 95\% confidence interval across 5 runs for \textbf{GRPO} (vanilla GRPO), \textbf{URS} (GRPO with random sampling $p=0.5$), \textbf{Det. Trunc.} (GRPO with deterministic prefix truncation of 50\% of trajectory tokens) and \textbf{RPC} (GRPO with uniform random prefix cutting). }
  \label{fig:grad_nrom}
\end{figure*}

\begin{figure*}
  \centering
  \begin{subfigure}[t]{0.4\columnwidth}
    \centering
    \includegraphics[width=\linewidth]{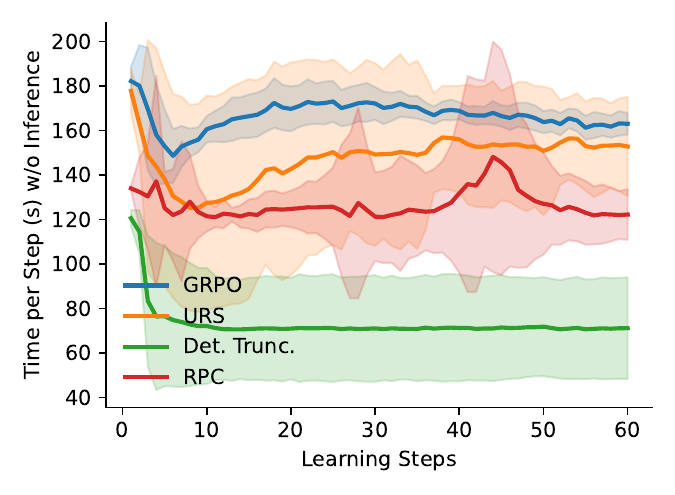}
    \caption{Qwen2.5-Math-7B}
    \label{fig:qwen_2_5_update_time}
  \end{subfigure}
  \hfill
  \begin{subfigure}[t]{0.4\columnwidth}
    \centering
    \includegraphics[width=\linewidth]{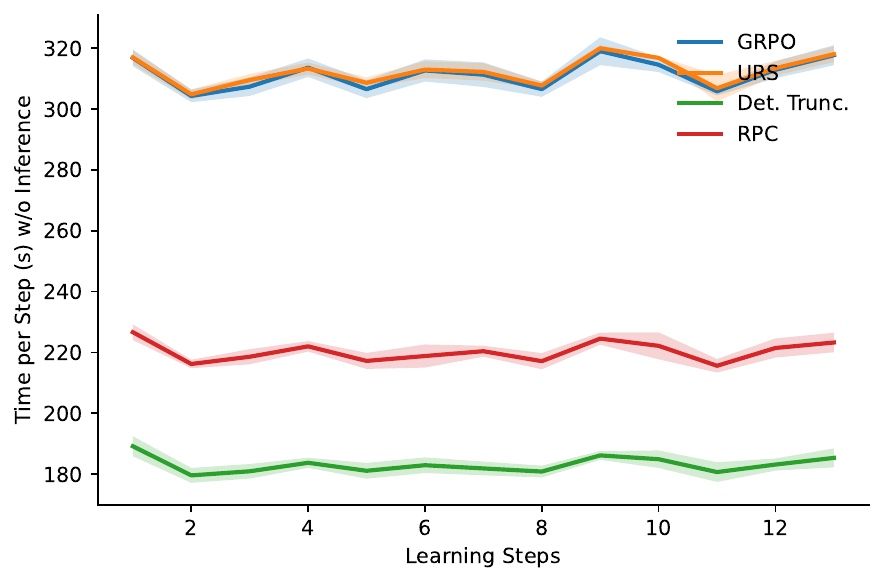}
    \caption{Qwen3-8B}
    \label{fig:qwen_3_update_time}
  \end{subfigure}
 \caption{
      Time per step (excluding inference time) with 95\% confidence interval across 5 runs for \textbf{RPC} (GRPO with uniform random prefix cutting). }
  \label{fig:update_time}
\end{figure*}

\begin{figure*}
  \centering
  \begin{subfigure}[t]{0.4\columnwidth}
    \centering
    \includegraphics[width=\linewidth]{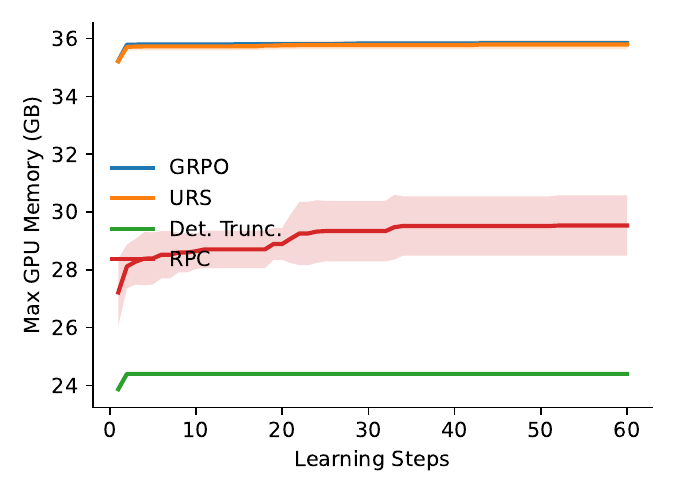}
    \caption{Qwen2.5-Math-7B}
    \label{fig:qwen_2_5_memory}
  \end{subfigure}
  \hfill
  \begin{subfigure}[t]{0.4\columnwidth}
    \centering
    \includegraphics[width=\linewidth]{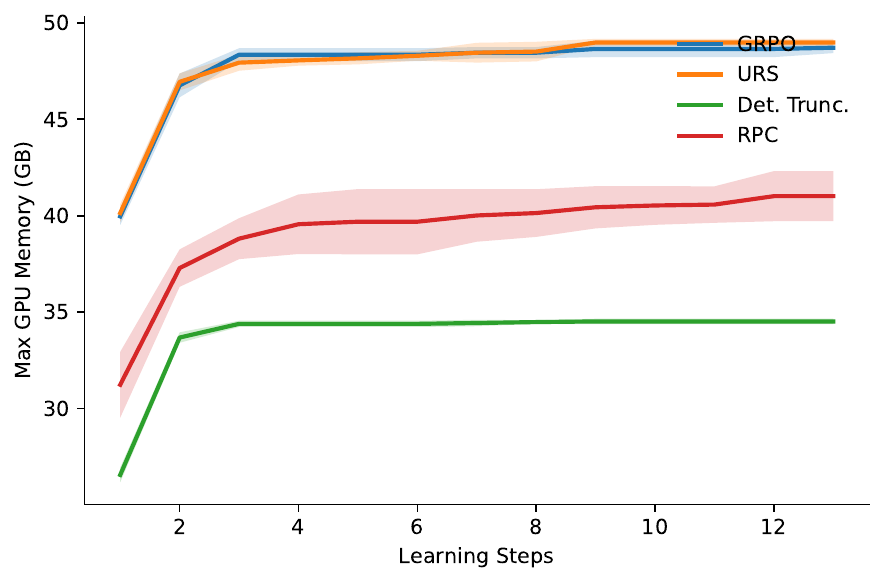}
    \caption{Qwen3-8B}
    \label{fig:qwen_3_memory}
  \end{subfigure}
 \caption{
      Max allocated GPU memory curves in \textbf{GB} with 95\% confidence interval across 5 runs for \textbf{GRPO} (vanilla GRPO), \textbf{URS} (GRPO with random sampling $p=0.5$), \textbf{Det. Trunc.} (GRPO with deterministic prefix truncation of 50\% of trajectory tokens) and \textbf{RPC} (GRPO with uniform random prefix cutting). }
  \label{fig:gpu_memory}
\end{figure*}